\documentclass[conference]{IEEEtran}

\usepackage[utf8]{inputenc}
\usepackage{graphicx}
\usepackage{float}
\usepackage{multirow}  
\usepackage{amssymb}
\usepackage{amsmath}

\usepackage[colorlinks=true, citecolor=blue,linkcolor=blue,bookmarks=false]{hyperref}
\usepackage[labelsep=quad,indention=10pt]{subfig}

\DeclareMathOperator*{\argmax}{arg\,max}
\newcommand{\loss}{\text{loss}}

\usepackage{adjustbox}
\usepackage[style=ieee]{biblatex} 
\bibliography{references}

\title{Robust SleepNets}

\author{
\IEEEauthorblockN{Yigit Alparslan}
\IEEEauthorblockA{\textit{Department of Computer Science} \\
\textit{Drexel University}\\
Philadelphia, PA, US \\
ya332@drexel.edu}
\and
\IEEEauthorblockN{Edward Kim}
\IEEEauthorblockA{\textit{Department of Computer Science} \\
\textit{Drexel University}\\
Philadelphia, PA, US \\
ek826@drexel.edu}
}

\begin{document}

\maketitle

\begin{abstract}
State-of-the-art convolutional neural networks excel in machine learning tasks such as face recognition, and object classification but suffer significantly when adversarial attacks are present. It is crucial that machine critical systems, where machine learning models are deployed, utilize robust models to handle a wide range of variability in the real world and malicious actors that may use adversarial attacks. In this study, we investigate eye closedness detection to prevent vehicle accidents related to driver disengagements and driver drowsiness. Specifically, we focus on adversarial attacks in this application domain, but emphasize that the methodology can be applied to many other domains. We develop two models to detect eye closedness: first model on eye images and a second model on face images. We adversarially attack the models with Projected Gradient Descent, Fast Gradient Sign and DeepFool methods and report adversarial success rate. We also study the effect of training data augmentation. Finally, we adversarially train the same models on perturbed images and report the success rate for the defense against these attacks. We hope our study sets up the work to prevent potential vehicle accidents by capturing drivers' face images and alerting them in case driver's eyes are closed due to drowsiness. 
\end{abstract}


\let\thefootnote\relax\footnote{{All code is open-sourced on \href{https://github.com/ya332/robust_sleepnets}{GitHub.}}}

\section{Introduction}
\label{sec:introduction}
Recent machine learning breakthroughs help solve many tasks including facial recognition \cite{aajfacialadversarial}, surveillance \cite{aplkc20}, natural language processing tasks \cite{daelemans2002evaluation}, materials discovery \cite{edwardkimmaterialsdiscovery} and bio-authentication systems \cite{deng2019arcface}. One of these tasks where machine learning models can be utilized is to detect whether a driver is drowsy inside the car and help prevent accidents with an alert system. National Highway Traffic Safety Administration (NHTSA) data shows 37,461 people were killed in 34,436 motor vehicle crashes, an average of 102 per day in 2016. Of all these accidents, alcohol-impaired driving fatal crashes totaled 9477 in 2016 \cite{nhtsa2016}. A potential alert system that detects driver disengagements while driving could prevent these accidents where the cause was alcohol or drowsiness. An estimated 1 in 25 adult drivers (aged 18 or older) report having fallen asleep while driving in the previous 30 days \cite{drowsydriving1} \cite{drowsydriving2}.
Additionally, drowsy driving was responsible for 72,000 crashes, 44,000 injuries, and 800 deaths according to the National Highway Traffic Safety Administration (NHTSA)  in 2013 \cite{drowsydriving3}. NHTSA also reports that up to 6,000 fatal crashes each year may be caused by drowsy drivers \cite{drowsydriving4} \cite{drowsydriving5} \cite{drowsydriving6}. Moreover, driving after going more than 20 hours without sleep is the equivalent of driving with a blood-alcohol concentration of 0.08\%, which is the U.S. legal limit. A person is three times more likely to be in a car crash if fatigued \cite{drowsydriving1}.
 Many car manufacturers incorporate assisted driving or drowsiness alert systems to their cars. Such cars may have an eye tracking camera built-in that records driver's face while driver is on the wheel. Other form of such alerting systems currently implemented to production level cars by car manufacturers may include pressure sensors on the steering wheel. Pressure sensors can be used to measure any long duration where the driver's hands are not on the steering wheel. In the event that this scenario happens, a car's alerting system would alert the driver via sound, or apply emergency brakes in case the driver is sleepy, not conscious or passed out due to a health incident. Currently such implementations exist for high-end luxury cars \cite{mercedes}.  Even though such steering interventions exist for high-end cars, the average car doesn't have such sophisticated drowsiness detection systems or intervention mechanisms. A general solution scalable to not only high-end luxury cars but also to general public cars could be a mobile application which could detect when driver closes their eyes for a certain amount of time and alert the driver via sound. 
 
A previous work \cite{Alparslan2020} by the authors resulted in an initial prototype published as a mobile app for Android users. See  \href{https://play.google.com/store/apps/details?id=driveup.facedetector}{here} for a working prototype of the said alerting system deployed and published as a mobile application. 
 Such mobile application would utilize a state-of-the-art neural network trained on images of people with closed and open eyes. Due to the nature of the task, it is crucial that such a neural network is agnostic to poor light conditions, out of zoom image captures, blurs, shadows, low resolution images, etc. Therefore, it is essential to have a robust model.

The title of this study, ``Robust SleepNets'', come from the idea of being able to detect fatigued or drowsy drivers (or any driver that might disengaged with driving for long duration) at the steering wheel. In this study, we assume eye closedness is a direct indicator of drowsiness so we focus on implementing neural network architectures that would detect eye closedness. We attack the models with adversarial attacks to study the robustness of the networks under adversarial conditions.  The robustness acts as a proxy to real life conditions which might exhibit poor lightning, out-of-focus camera zoom, overexposed/underexposed shutter, height or width shift in the frame etc. We also investigate augmenting train data with the parameters described in \autoref{tab:data_augmentation_training_parameters} to study the impact of data augmentation on accuracy and adversarial defense. 

In this study, we assume the following two problems are functionally equivalent. 
\begin{enumerate}
    \item Detecting drowsy driving 
    \item Detecting whether driver closes their eyes. (longer than some threshold that could be determined empirically. such threshold is not the focus of this study)
\end{enumerate}

In other words, we focus on detecting eye closedness to detect drowsy driving. Then, we \textbf{treat the real-word driving conditions such as poor light, out-of-focus camera etc as black-box adversarial attacks}. With these assumptions and motivations, we investigate the usage of adversarial training as a means to creating robust models that could detect eye closedness against said adversarial attacks. Additionally, we study the effect of training data augmentation.  

This research is organized so that \autoref{sec:introduction} introduces the concept of drowsy driving and \autoref{sec:relatedwork} explores what has been done in this field. \autoref{sec:methodology} explains the dataset, the models, adversarial attacks, adversarial training and limitations that we use/have in this study. We report the results in \autoref{sec:results} and conclude the study in \autoref{sec:conclusion} and \autoref{sec:futurework} with summarizing what we have done in this study and discussing where the research might go in the future.

\section{Related Work}
\label{sec:relatedwork}
There has been previous results where eye detection was used as a gateway to detect drowsiness \cite{chen2017}. There has been little study to investigate the robustness of these models with adversarial attacks. 

Adversarial attacks are inputs that look like the original images but with perturbations added to result in misclassifications in the classifier \cite{AdvAttacks} \cite{kim2020modeling} \cite{edwardkimregularization} \cite{kim2018deep}. Adversarial attacks can be created in the image domain \cite{aajfacialadversarial} as well as audio domain \cite{aab20audio}.

Adversarial training is one way of defending against these attacks since using adversarial attacks \cite{madry2017towards}. we can generate adversarial samples and then use these samples in our training to develop high accuracy models. Adversarial training as a defense depends on model and task at hand significantly.   

Alparslan et al. \cite{Alparslan2020} investigated the eye and face models on driver detection and explored data augmentation to simulate real-world black-box adversarial image settings.  Even though they did not apply adversarial attacks, or adversarial training, they claimed that adding a robust and systematic data augmentation to the training datasets would represent black-box attacks in a real-world scenario where driver face might be in too much light, or shadow, or it might be blurred if the camera angle is not adjusted.  

In this study, we include data augmentation that Alparslan et al. included in their study. In addition to the data augmentation, we also investigate adversarial robustness by attacking the models with Projected Gradient Descent, Fast Gradient Sign and DeepFool attacks. Once we report the accuracy on the adversarially generated dataset, we feed the adversarial inputs back into the training dataset to defend against the attacks to study the possibility of creating robust models that would detect eye closedness and drowsiness in the presence of attacks.

\section{Methodology}
\label{sec:methodology}
We develop two models: first on eye images and second on face images. We use Eye-blink dataset \cite{Pan2007EyeblinkbasedAI} for the eye model and we use Closed-Eye in the Wild dataset \cite{SONG20142825} for the face model. 

For the eye model, we use 3,108 images belonging to 2 classes (open and closed) to train and we use 776 images to cross-validate during training. We use 962 images that the model has never seen before to just test the data. For the eye model, all classification results in \autoref{table:all_results} are reported from these 962 images of test dataset.

For face model, we use 1,559 images belonging to 2 classes (open and closed) to train and we use 389  images to cross-validate during training. We use 485 images that the model has never seen before to just test the data. For the face model, all classification results in \autoref{table:all_results} are reported from these 485 images of test dataset.

Additionally, in this study, we investigate the possible impact of data augmentation against adversarial attacks. For both models, we report results for the model trained on augmented data as well as non-augmented data. The data augmentations that we apply are summarized below:

\begin{enumerate}
\label{list:data_augmentation}
    \item \textbf{Rotation:} Image can be rotated randomly depending on the driver's position and the camera angle.
    \item \textbf{Width Shift:} Image width might depend on the camera angle. The model needs to mitigate this randomness .
    \item \textbf{Height Shift:} Image height might depend on the camera angle. The model needs to mitigate this randomness.
    \item \textbf{Shear angle Shift:} Drivers plane intersects with the plane in which camera is mounted on a car. This creates additional randomness and the model needs to mitigate this randomness.
    \item \textbf{Zoom:} The camera can be close or far to the driver and the model needs to mitigate this randomness.
    \item \textbf{Horizontal Flip:} This doesn't correlate to real life setting, but the idea is that the driver's window will be always at its left side, which means the lightning conditions from the left side of the camera will be always poor compared to the right side. The model needs to detect fatigue regardless of the lightning hence the flip.
    \item \textbf{Image Fill:} Image can be scaled down or up. The model needs to detect fatigue regardless.
    \item \textbf{Scaling:} Image can be scaled down or up. The model needs to detect fatigue regardless.
\end{enumerate}

Additionally, we apply Projected Gradient Descent \cite{madry2017towards}, Fast gradient Sign Method \cite{goodfellow2014explaining} and DeepFool \cite{moosavi2015deepfool} to attack the models and report their accuracies.

\subsection{Adversarial Attacks}
\subsubsection{PGD}
\label{attack:pgd}
Projected Gradient Descent~\cite{madry2017towards} is a strategy for finding an adversarial example $x'$ for an input $x$ that satisfies a given norm-bound $\|x' -x\|_p \leq \epsilon$.

Let $B$ denote the $\ell_p$-ball of radius $\epsilon$ centered at $x$.
The attack starts at a random point $x_0 \in B$, and repeatedly sets
\begin{align*}
x_{i+1} &= \text{Proj}_{B}(x_i + \alpha \cdot g) \\
\quad \text{for } g &= \argmax_{\|v\|_p \leq 1} v^\top \nabla_{x_i} L(x_i, y) \;.
\end{align*}
Here, $L(x, y)$ is a suitable loss-function (e.g., cross-entropy), $\alpha$ is a step-size, $\texttt{Proj}_B$ projects an input onto the norm-ball $B$, and $g$ is the \emph{steepest ascent} direction for a given $\ell_p$-norm. E.g., for the $\ell_\infty$-norm, $\texttt{Proj}(z)$ is a clipping operator and $g = \texttt{sign}(\nabla_{x_i} L(x_i, y))$ .

\subsubsection{Fast Gradient Sign}
\label{attack:fgs}
The fast gradient sign \cite{goodfellow2014explaining} method optimizes for the $L_{\infty}$ distance metric and its advantage is fast running tim, which comes at the expense of generating images that are very similar to the original image.

Given an image $x$ the fast gradient sign method sets
\begin{equation*}
  x' = x - \epsilon \cdot \text{sign}(\nabla \loss_{F,t}(x)),
\end{equation*}
where $\epsilon$ is chosen to be sufficiently small so as to be undetectable,
and $t$ is the target label. 
Intuitively, for each pixel, the fast gradient sign method uses the gradient
of the loss function to determine in which direction the pixel's intensity should be changed (whether it should be increased or decreased) to minimize the loss function; then, it shifts all pixels
simultaneously.

\subsubsection{DeepFool}

Deepfool \cite{moosavi2015deepfool} optimizes over $L_2$ distance metric with the assumption that neural networks are linear. Since neural networks are not linear, once a hyperplane is found (if found any) that seperates two classes, search terminates. DeepFool takes about 10x more than PGD and FGSM takes on average when we apply for our eye and face models.  
We refer the authors to the work of Moosavi et al. \cite{moosavi2015deepfool} for more in-depth explanation.

\subsection{Adversarial Training}
Adversarial training \cite{fgm} \cite{kannan2018adversarial} \cite{madry2017towards} is the process where the adversarially generated samples are included in the training data in the hopes that the model will recognize the attacks next time sees it. In the current literature, adversarial training is one of the stronger defenses against adversarial attacks especially if it is combined with other defenses \cite{madry2017towards} \cite{tramer2020adaptive}. We adversarially train our augmented eye model, non-augmented eye model, augmented face model and finally non-augmented face model. During our adversarial training, we attack entirety of the training dataset via Projected Gradient Descent, Fast Gradient Sign Method and DeepFool attacks. Since we use one attack to adversarially train one model and there are three attacks, we report three adversarially trained models for the Eye-blink dataset and three adversarially trained models for the Closed Eyes in the Wild dataset.  We repeat the entire process once for the augmented data case and once for non-augmented data case, which doubles our combinations.

\subsection{Data Augmentation}
Data augmentation is the process of generating new training data from the existing samples. Some data augmentation methods include adding rotation, adding text, adding zoom, height shift, width shift.  
In this study, we also investigate the effect of data augmentation in the presence of adversarial attacks. The data augmentation parameters are the same for both eye and face model and can be seen in \autoref{tab:data_augmentation_training_parameters}. These parameters are the same as the parameters described in \cite{Alparslan2020}

\begin{table}[!htbp]
\caption{Noisified training data parameters are shown below. Adding noise to the training data to augment it simulates black-box settings in real-world scenarios where  a driver face might blurred, occluded, underexposed/overexposed etc. In these scenarios, the data augmentation helps simulate an adversary under black box settings.}
\label{tab:data_augmentation_training_parameters}
\centering
 \begin{tabular}{|c|c|} 
 \hline
 \textbf{Change Type} & \textbf{Change Amount} \# \\ [0.5ex] 
 \hline
 rotation range & 40$^o$\\ 
 \hline
  width shift range  & 0.2 \\
 \hline
 height shift range & 0.2 \\
 \hline
 shear range & 0.2\\
 \hline
 zoom range & 0.2\\ 
 \hline
 horizontal flip & True\\ 
 \hline
 fill mode & 'nearest' \\ 
 \hline
 rescale &  1./255\\ 
 \hline
\end{tabular}
\end{table}

\subsection{Models}
\begin{figure}[htbp!]%
    \centering
    \subfloat[Eye model accuracy for training and testing dataset versus epoch count ]{\includegraphics[scale=0.5]{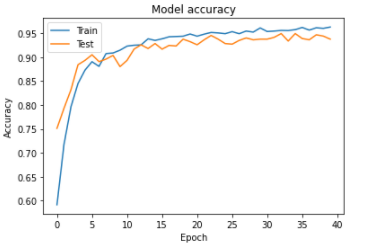}}
    \qquad
    \subfloat[Eye model binary cross entropy loss for training and testing datasets versus epoch count]{\includegraphics[scale=0.5]{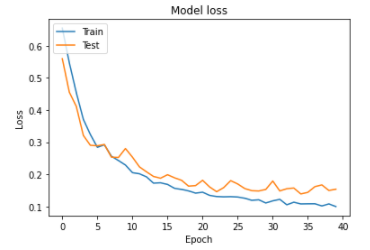} }%
    \caption{Training and testing of eye model with Eye-blink dataset when training data is \textbf{non-augmented}.}
    \label{fig:accuracy_and_loss_nonaugmented_eye}
\end{figure}

\begin{figure}[htbp!]%
    \centering
    \subfloat[Eye model accuracy for training and testing dataset versus epoch count ]{\includegraphics[scale=0.5]{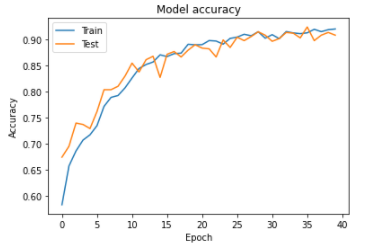}}
    \qquad
    \subfloat[Eye model binary cross entropy loss for training and testing datasets versus epoch count when training data is \textbf{augmented}]{\includegraphics[scale=0.5]{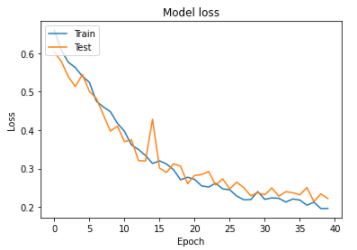} }%
    \caption{Training and testing of eye model with Eye-blink dataset when training data is \textbf{augmented}.}
    \label{fig:accuracy_and_loss_augmented_eye}
\end{figure}

We use the same model architecture for the Eye-blink dataset and the Closed Eyes in the Wild dataset in order to eliminate the differences in accuracy that might arise due to architecture configurations. Since both dataset represents a binary classification problem, we argue that using the same architecture for both datasets doesn't cause problem to detect eye closedness. 
For eye model, we use 3,108 images belonging to 2 classes as training data and 776  images belonging to 2 classes as validation data to train and cross validate our models (See \autoref{fig:accuracy_and_loss_augmented_eye}, \autoref{fig:accuracy_and_loss_nonaugmented_eye} to see the training accuracy and loss plots). Then, we report the testing accuracy on the testing dataset 962 images belonging to 2 classes (All numbers are from testing accuracy in \autoref{table:all_results}). The test dataset consist of images that are never seen before by the model.

\begin{figure}[!htbp]%
    \centering
    \subfloat[Face model accuracy for training and testing dataset versus epoch count ]{\includegraphics[scale=0.5]{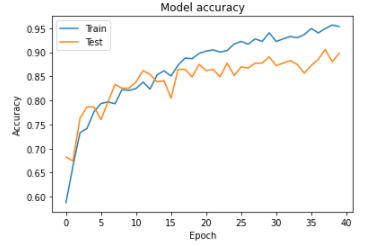}}
    \qquad
    \subfloat[Face model binary cross entropy loss for training and testing datasets versus epoch count]{\includegraphics[scale=0.5]{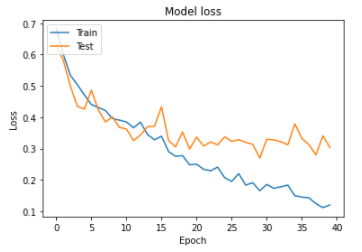} }%
    \caption{Training and testing of face model with CEW dataset when training data is \textbf{non-augmented}.}
    \label{fig:accuracy_and_loss_nonaugmented_face}
\end{figure}

\begin{figure}[!htbp]%
    \centering
    \subfloat[Face model accuracy for training and testing dataset versus epoch count ]{\includegraphics[scale=0.5]{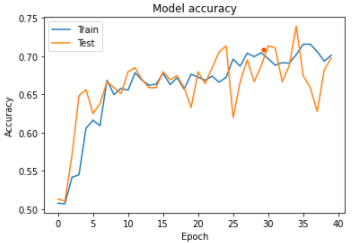}}
    \qquad
    \subfloat[Face model binary cross entropy loss for training and testing datasets versus epoch count]{\includegraphics[scale=0.5]{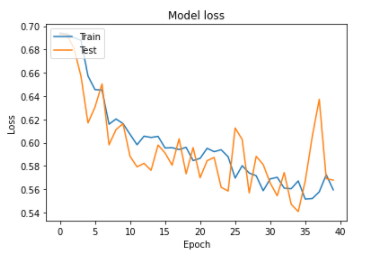} }%
    \caption{Training and testing of face model with CEW dataset when training data is \textbf{augmented}.}
    \label{fig:accuracy_and_loss_augmented_face}
\end{figure}

For face model, we use 1,559 images belonging to 2 classes as training data and 389 images belonging to 2 classes as validation data to train and cross validate our models (See \autoref{fig:accuracy_and_loss_augmented_face},
\autoref{fig:accuracy_and_loss_nonaugmented_face} to see the training accuracy and loss plots). Then, we report the testing accuracy on the testing dataset 485 images belonging to 2 classes (All numbers are from testing accuracy in \autoref{table:all_results}). The test dataset consist of images that are never seen before by the model. 

Model architectures can be seen in \autoref{tab:model_architecture}.

\begin{table}[!htbp]
\caption{Model Architecture for both models. We used the same architecture on the Eye-Blink and CWE dataset. The eye model and the face model both use binary cross entropy with Adam's optimizer on iterative gradient descent.}
\vspace{2em}
\label{tab:model_architecture}
\centering
 \begin{tabular}{|c|c| c|} 
 \hline
 \textbf{Layer type} & \textbf{Output Shape} & \textbf{Param} \# \\ [0.5ex] 
 \hline
 Conv2D & (98, 98,6) & 60 \\ 
 \hline
 Average Pooling  & (49,49,6) & 0 \\
 \hline
 Conv2D & (47,47,16) & 880 \\
 \hline
 Average Pooling & (23,23,16) & 0  \\
 \hline
 Flatten & 8464 & 0 \\ 
 \hline
 Dense & 120 & 1015800 \\ 
 \hline
 Dense & 84 & 10164 \\ 
 \hline
 Dense & 1 & 85 \\ 
 \hline
\end{tabular}
\end{table}

\subsection{Dataset}

In this study, we use Eye-blink Dataset \cite{f14} data set to  train our eye model. Eye Blink dataset has 2,100 closed and open eye images that are black and white. They are 24x24 pixels and only show eye patches. We use Closed Eyes In The Wild (CEW) \cite{f14} dataset to train our face model. CEW dataset has 1,192 subjects with both eyes closed and 1,231 subjects with eyes open. Some challenges of this set include amateur photography, occlusions, problematic lighting, pose, and motion blur.

\begin{figure}
\centering
  \includegraphics[width=0.9\columnwidth]{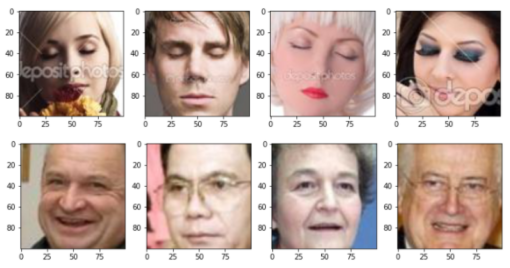}
  \caption{Samples from Closed Eyes in the Wild dataset. This dataset includes 1,231 opened-eye images of people and 1192 closed-eye images of people. Some difficulties with this dataset include blur, fade, shade and over/underexposing. Due to these innate difficulties, this dataset represents well the environment an actual driver might be in while driving in a real-world scenario.}~\label{fig:cew_dataset}
\end{figure}

\begin{figure}
\centering
  \includegraphics[width=0.9\columnwidth]{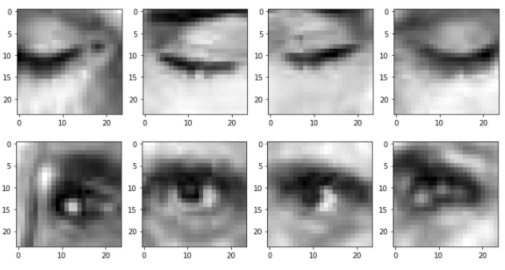}
  \caption{Samples from Eye-blink dataset. This dataset consists 2100 closed and open 24x24 pixel grayscale eye patch images. Some difficulties with this dataset include very low resolution and shade.}~\label{fig:figure2}
\end{figure}

\subsection{Limitations}
In this study, we examine many different combinations to see the full effect of attacks and defenses. We have two models and for each model, we repeat the experiments for the case when the training data is augmented and for the case where it is not augmented.  We use three attacks on each model and also apply adversarial training for each of the attacks. Because we have about 16 different configurations when all combined in this study, we are generating grayscale adversarial images even though the input can be colored (applying the attack for each channel). This helps us speed up the training, attacking and defending duration. The readers for example will see from the \autoref{table:all_results} that PGD doesn't defend successfully when attacked. Researchers who might reproduce our combinations in the future might get different results if this nuance is not taken into consideration.

\section{Experiment Results and Evaluation}
\label{sec:results}

We report all the results in \autoref{table:all_results}. In our analysis, we need to consider two aspects of the attacks. 
\begin{itemize}
    \item How successful is the adversarial attack? 
    \item How successful is the defense?
\end{itemize}

In order to answer the first part, (i.e how successful the attacks are), we attack the entire test data and report how accurate the model is against them. The accuracies for this case are represented in parenthesis in the \autoref{table:all_results}. For example, for the case of Eye model \textbf{without} the data augmentation, PGD attack reports 81.70\% accuracy. This means the model is able to accurately assign a class label in the images adversarially altered via PGD attack, 81.70\% of the time. Since this accuracy value is fairly high, we conclude that PGD attack was not able to fool the classifier and failed as an \emph{attack}. The same model is then adversarially trained with the images that were altered via PGD attack, and reports 81.70\% accuracy, which suggests that the adversarial training doesn't have any success as a defense at all. Readers might be curious why the same accuracy is reported (81.70\% vs 81.70\%). This is quite possible if adversarial training is an identity function. (i.e adversarial training is not a successful defense since model reports the same accuracy for the attacks before the adversarial training and after the adversarial training). 

A high accuracy value in the accuracy column of the \autoref{table:all_results} does not signify any conclusion regarding the success of the adversarial training. Mainly, the increase from the accuracy value inside the parenthesis tells how successful the adversarial training was in \autoref{table:all_results}. To answer the second part (i,e how successful the defenses), we calculate and compare that difference.

\subsubsection{PGD}
All PGD rows reports very high accuracy values ($> 81\%$) for all models after the adversarial training, but this does \emph{not} mean the adversarial training was successful since this can be attributed to the fact that PGD attack cannot lower the accuracy of the models to begin with. (All accuracies inside the parenthesis for the PGD rows are ($>79\%$). We conclude PGD cannot fool the classifier successfully. So, adversarial training via PGD attack doesn't offer significant improvements. On average PGD attack decreases the classifier accuracy 9.44\% for the eye model and 0.1\% for the face model.  

On average PGD defense increases the classifier accuracy -1.2\% for the eye model and 0.2\% for the face model (Please note -1.2\% increase means 1.2\% decrease). This can be empirically checked via the samples generated by PGD in \autoref{sec:appendix} since adversarially generated samples that used PGD method look very similar to the original images. PGD attack/defense failure can be attributed to the fact that eye closedness detection is a very localized task on an image and PGD can't successfully alter such information.

\subsubsection{FGSM and DeepFool}
In the case of FGSM and DeepFool, as it can be seen in \autoref{table:all_results}, these two attacks can lower the accuracy of the models significantly compared to the baseline. (around 43\% ) 

On average FGSM attack decreases the classifier accuracy 42.24\% for the eye model and 32.22\% for the face model. On average FGSM defense increases the classifier accuracy 1.1\% for the eye model and 0\% for the face model.

On average DeepFool attack decreases the classifier accuracy 31.43\% for the eye model and 28.77\% for the face model.  On average DeepFool defense increases the classifier accuracy -9.81\% for the eye model and -23.5\% for the face model. (Please note a -9.81\% increase means 9.81\% decrease so adversarially training a model via DeepFool attack actually worked the opposite way of its intent.)

For these attacks, however, we conclude the defense is not successful since adversarially trained models cannot produce accuracy values greater than the accuracy values on the adversarial attacks before the training. We conclude for FGSM and DeepFool that the attacks are extremely strong and defenses are not. We invite the reader to examine the adversarially generated samples in \autoref{sec:appendix}.

\subsubsection{Data augmentation}
For the face model, accuracy drops 17\% when training data is augmented compared to the case when training data is \emph{not} augmented.

However, we also see that data augmentation helps the classifier become more robust. On average for all the attacks, when training data is augmented, classifier reports 1.3\% higher accuracy when attacked. 

For the eye model, accuracy drops 2\% when training data is augmented compared to the case when training data is \emph{not} augmented.

However, we also see that data augmentation helps the classifier become more robust. On average for all the attacks, when training data is augmented, classifier reports 6.14\% higher accuracy when attacked. 

\subsubsection{Eye vs Face Model}
Accuracy values indicate that eye closedness detection is easier to achieve when done on only eye images. On average, eye models classify 12\% more accurately than face model for the base cases. This can be attributed to the fact that face image has redundant information such as hair, mouth, ears etc (i.e all part of the face other than eyes) that the classifier needs to learn whereas eye model does not have such overhead. Because eye detection is such localized task, this could also explain why data augmentation helps the eye model 5x more than the face model when attacked. Defending the area of eyes via augmentation is easier on the eye images than face images.

\begin{table}[htbp!]
\centering
\caption{Accuracy results for the adversarially trained eye and face model. The third column (Config) reports the configuration of the attack type used during adversarial training. The accuracy column represents accuracy on newly generated adversarial samples after adversarial training. The increase compared to the accuracy value inside the parenthesis tells how successful the adversarial training was. Inside the parenthesis, it reports accuracy on original adversarial samples, which are generated on the model before its adversarial training. A high accuracy value inside the parenthesis means the attack couldn't lower the accuracy successfully. Overall, the eye model has performed better than the face model and data augmentation reduces accuracies more for the face model than eye model.  }
 \begin{adjustbox}{max width=\columnwidth}
\begin{tabular}{cc|c|c|c|c|c|}
\hline
\multicolumn{1}{|c|}{\textbf{Model}} &
  \textbf{D.A} &\textbf{  Config  } & \textbf{Accuracy} & \textbf{Precision} & \textbf{Recall} & \textbf{F-1 Score} \\ \hline

\multicolumn{1}{|c|}{\multirow{8}{*}{\textbf{Eye}}} &
{\multirow{4}{*}{\textbf{ W/O }}} & \textbf{Base} & 95 & 96 & 96 & 95.5\\
\multicolumn{1}{|c}{} & \multicolumn{1}{|c|}{} & \textbf{PGD} & 81.70 (81.70) & 84 & 82& 81\\ 
\multicolumn{1}{|c}{} & \multicolumn{1}{|c|}{} & \textbf{FGSM } & 54.47 (52.18) & 73 & 54& 42\\
\multicolumn{1}{|c}{} & \multicolumn{1}{|c|}{} & \textbf{DeepFool } & 55.09 (55.69) & 64 & 55& 45\\
\cline{2-7}
\multicolumn{1}{|c|}{} & {\multirow{4}{*}{\textbf{ W/ }}}  & \textbf{Base} & 93 & 92  & 94& 93\\ 
\multicolumn{1}{|c}{} & \multicolumn{1}{|c|}{} & \textbf{PGD} & 85 (87.42) & 85 & 85& 85\\ 
\multicolumn{1}{|c}{} & \multicolumn{1}{|c|}{} & \textbf{FGSM } & 51.14 (51.14) & 26 & 51& 35\\
\multicolumn{1}{|c}{} & \multicolumn{1}{|c|}{} & \textbf{DeepFool } & 50.42 (69.44) & 45 & 50& 36\\
\hline

\multicolumn{1}{|c|}{\multirow{8}{*}{\textbf{Face}}} &
{\multirow{4}{*}{\textbf{ W/O }}} & \textbf{Base} & 90 & 90 & 90 & 89\\
\multicolumn{1}{|c}{} & \multicolumn{1}{|c|}{} & \textbf{PGD} & 81.03 (79.18) & 81 & 81& 81\\ 
\multicolumn{1}{|c}{} & \multicolumn{1}{|c|}{} & \textbf{FGSM } & 49.28 (49.28) & 24 & 49& 33\\
\multicolumn{1}{|c}{} & \multicolumn{1}{|c|}{} & \textbf{DeepFool } & 52.58(62.47) & 53 & 53& 48\\
\cline{2-7}
\multicolumn{1}{|c|}{} & {\multirow{4}{*}{\textbf{ W/ }}}  & \textbf{Base} & 73 & 74 & 73& 73\\ 
\multicolumn{1}{|c}{} & \multicolumn{1}{|c|}{} & \textbf{PGD} & 82.68 (84.12)& 83 & 83& 83\\ 
\multicolumn{1}{|c}{} & \multicolumn{1}{|c|}{} & \textbf{FGSM } & 50.72 (50.72) & 26 & 51 & 34\\
\multicolumn{1}{|c}{} & \multicolumn{1}{|c|}{} & \textbf{DeepFool } & 49.90 (60.0) & 54 & 50 & 37\\
\hline
\end{tabular}
\end{adjustbox}
\label{table:all_results}
\end{table}

\section{Conclusion}
\label{sec:conclusion}
In this paper, we assumed eye closedness was a gateway to detecting driver fatigue. We trained two deep convolutional neural network models to detect eye closedness: one based on Eye-blink dataset and other based on Closed Eyes in the Wild dataset (CEW). Later, we crafted adversarial attacks via Fast Gradient Sign Method, Projected Gradient Descent Method and DeepFool attacks. Highest models for eye detection were baseline models without data augmentation with 95\% accuracy for the eye model and 90\% for face model. We conclude that PGD attack is not able to decrease classifier accuracy as much as FGSM and DeepFool attacks decrease (9.44\%, 42.24\%, 31.44\% decrease for eye model and  0.1\%, 32.22\%, -28.77\% decrease for the face model, respectively). Additionally, adversarially training the model with PGD attack or FGSM attack donot increase classifier accuracy. PGD reports 1.2\% (eye) and 0.2\% (face) accuracy decrease, FGSM reports 1.1\% (eye) and 0\% (face) accuracy increase and DeepFool reports 9.81\% (eye) and 23.5\% (face) accuracy \emph{decrease} which prove that DeepFool does not succeed at defending when used as adversarial training. Finally, eye model reports 6.14\% higher accuracy when it is trained on augmented data when attacked compared to the case where it is trained on non-augmented data. Face model reports 1.3\% higher accuracy when it is trained on augmented data  when attacked compared to the case where it is trained on non-augmented data. We hope that our robustness study help emphasize the need for robust machine learning models in mission-critical systems in the presence of adversarial attacks. We also hope that this study gives more insight on robust eye closedness detection methods as well as the effect of data augmentation and adversarial attacks as defense tools.

\section{Future Work}
\label{sec:futurework}
In the future, combining the two models in an ensemble learning setting would yield different results for a given image and might be worthwhile to examine.
Additionally, another future work might include improving the inference duration to enable eye closedness detection in real-time or on a video.

\printbibliography

\newpage
\section{Appendix}

\vspace{3.5cm}
\label{sec:appendix}
\begin{figure}[htbp!]
    \centering
    \subfloat[Eye-blink dataset non-augmented samples ]{\includegraphics[width=0.7\columnwidth, scale=0.4]{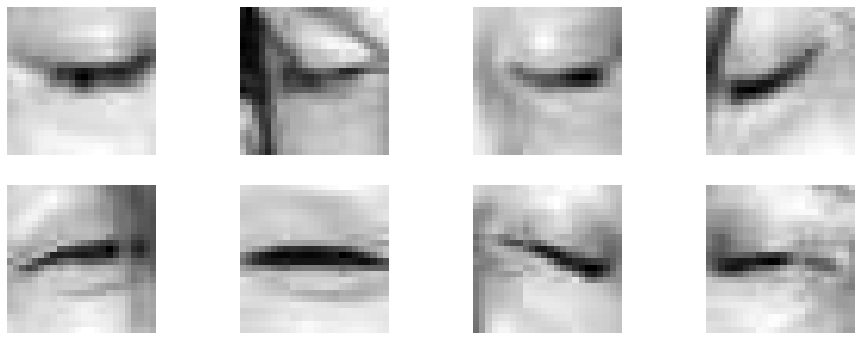}}
    \qquad
    \subfloat[Eye-blink dataset non-augmented samples after applying PGD]{\includegraphics[width=0.7\columnwidth, scale=0.4]{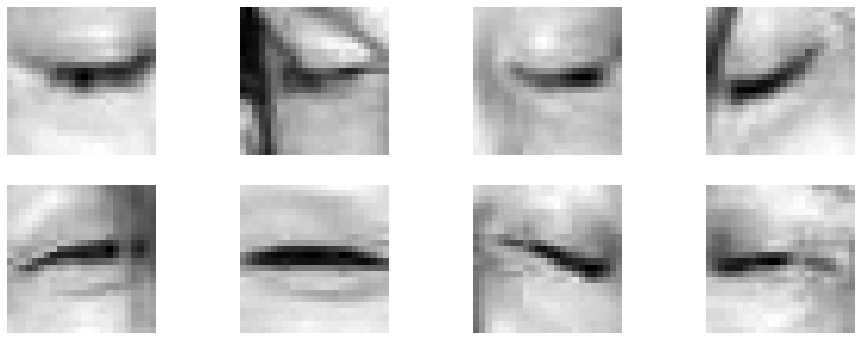} }%
    \qquad
    \subfloat[Eye-blink dataset non-augmented samples after applying FGSM ]{\includegraphics[width=0.7\columnwidth, scale=0.4]{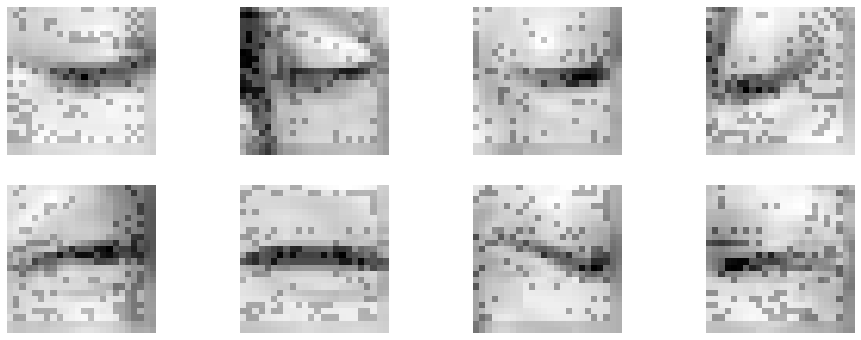}}
    \qquad
    \subfloat[Eye-blink dataset non-augmented samples after applying DeepFool ]{\includegraphics[width=0.7\columnwidth, scale=0.4]{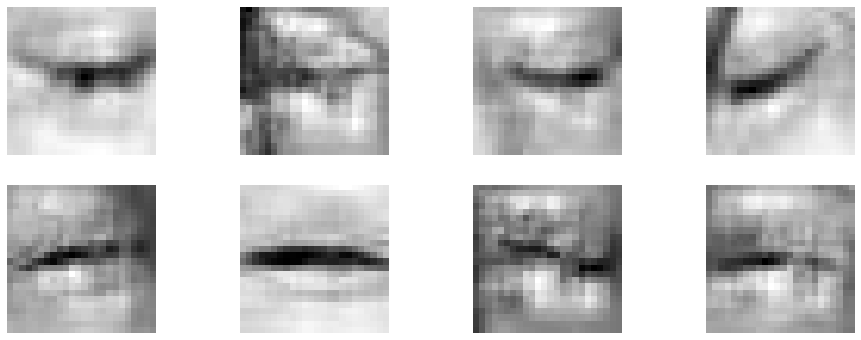}}
    \caption{Baseline images, FGSM, PGD and DeepFool adverserial attacks on non-augmented Eye-blink dataset.}
    \label{fig:eye_blink_without_augmentation}
\end{figure}

\begin{figure}[htbp!]
    \centering
    \subfloat[Eye-blink dataset augmented samples ]{\includegraphics[width=0.7\columnwidth, scale=0.4]{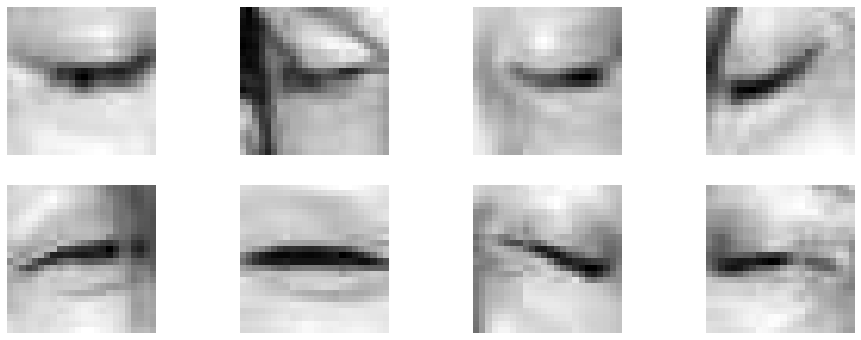}}
    \qquad
    \subfloat[Eye-blink dataset augmented samples after applying PGD]{\includegraphics[width=0.7\columnwidth, scale=0.4]{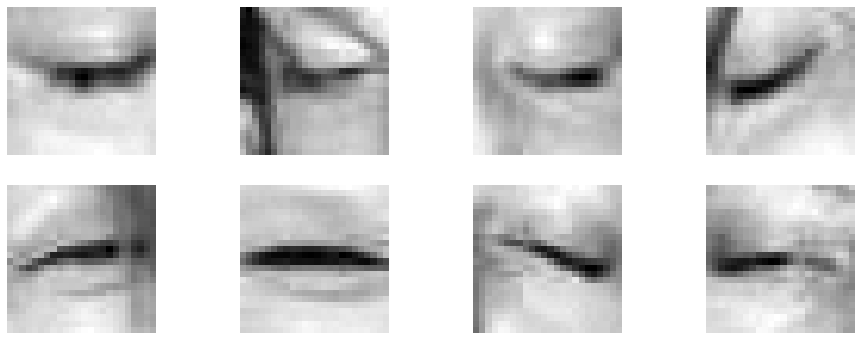} }%
    \qquad
    \subfloat[Eye-blink dataset augmented samples after applying FGSM ]{\includegraphics[width=0.7\columnwidth, scale=0.4]{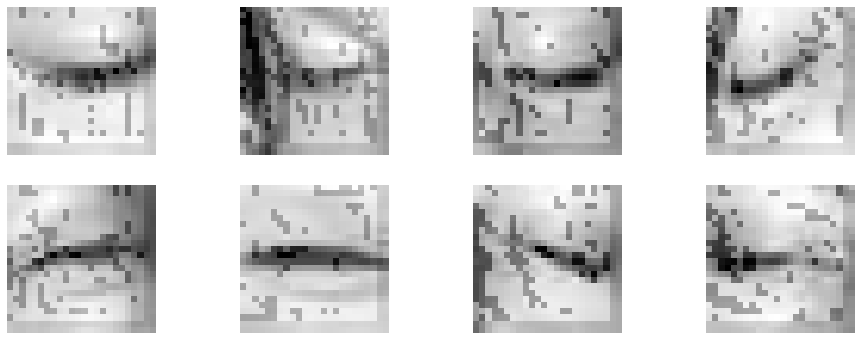}}
    \qquad
    \subfloat[Eye-blink dataset augmented samples after applying DeepFool ]{\includegraphics[width=0.7\columnwidth, scale=0.4]{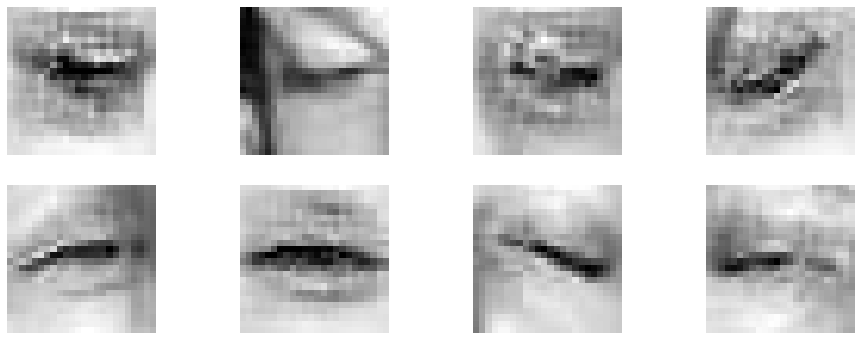}}
    \caption{Baseline images, FGSM, PGD and DeepFool adverserial attacks on augmented Eye-blink dataset.}
    \label{fig:eye_blink_with_augmentation}
\end{figure}

\begin{figure}[htbp!]
    \centering
    \subfloat[CEW dataset non-augmented samples ]{\includegraphics[width=0.7\columnwidth, scale=0.4]{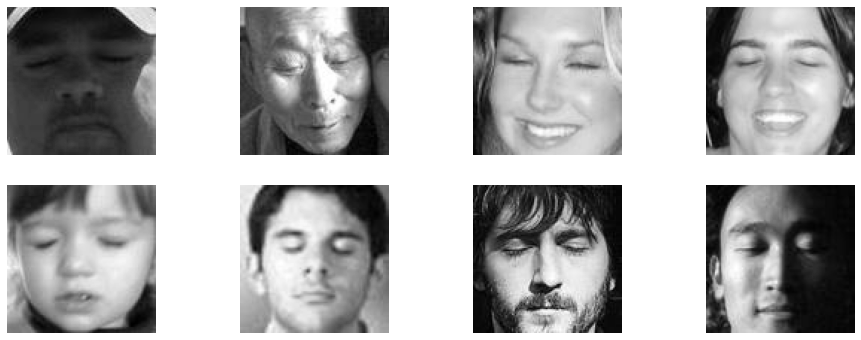}}
    \qquad
    \subfloat[CEW dataset non-augmented samples after applying PGD]{\includegraphics[width=0.7\columnwidth, scale=0.4]{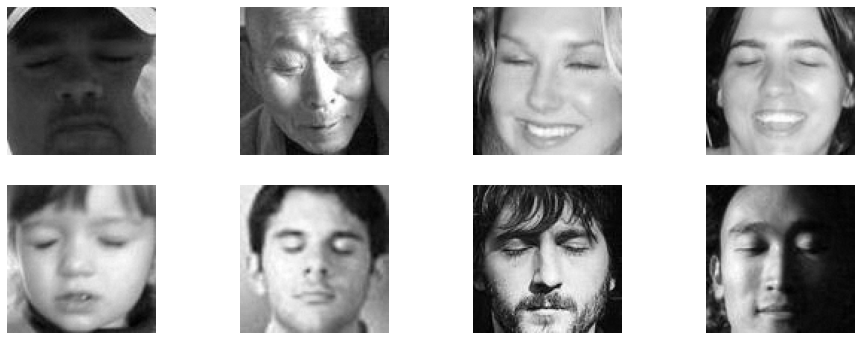} }%
    \qquad
    \subfloat[CEW dataset non-augmented samples after applying FGSM ]{\includegraphics[width=0.7\columnwidth, scale=0.4]{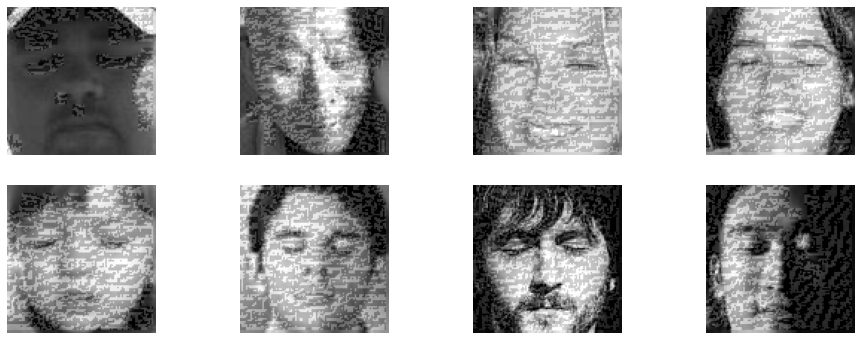}}
    \qquad
    \subfloat[CEW dataset non-augmented samples after applying DeepFool ]{\includegraphics[width=0.7\columnwidth, scale=0.4]{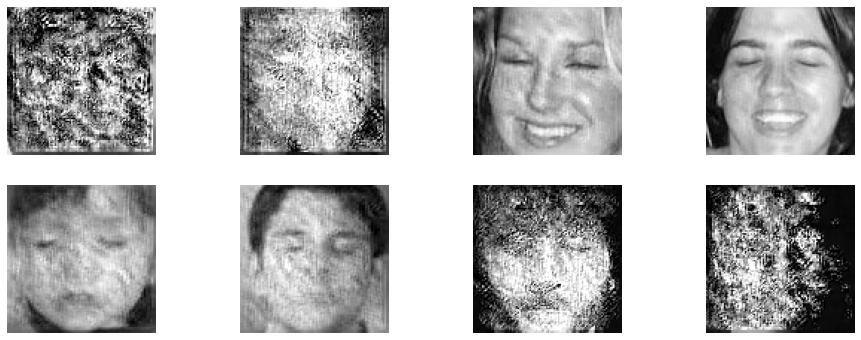}}
    \caption{Baseline images, FGSM, PGD and DeepFool adverserial attacks on non-augmented Closed Eyes in the Wild (CEW) dataset.}
    \label{fig:cew_without_augmentation}
\end{figure}

\begin{figure}[htbp!]
    \centering
    \subfloat[CEW dataset augmented samples ]{\includegraphics[width=0.7\columnwidth, scale=0.4]{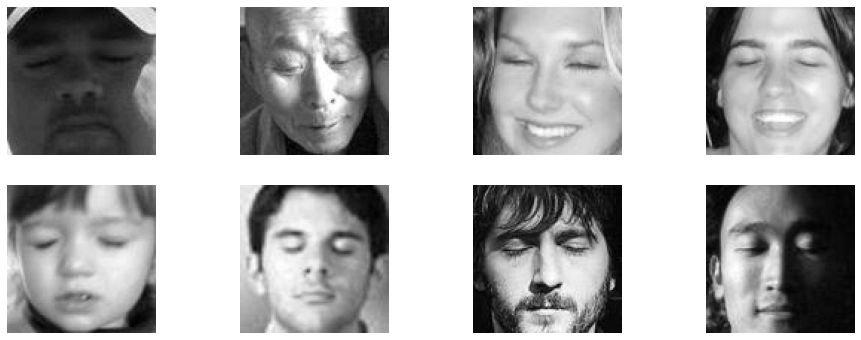}}
    \qquad
    \subfloat[CEW dataset augmented samples after applying PGD]{\includegraphics[width=0.7\columnwidth, scale=0.4]{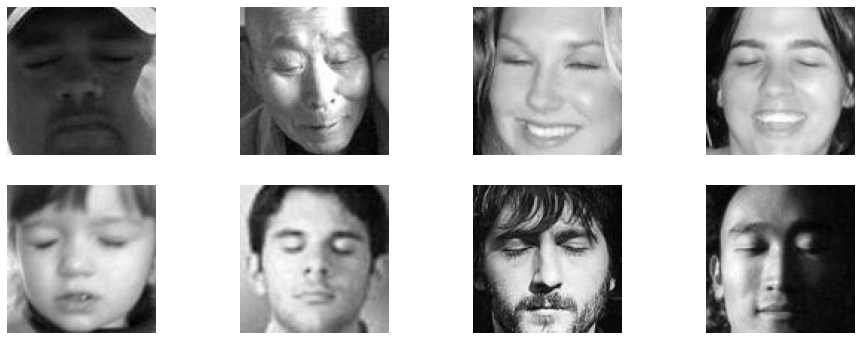} }%
    \qquad
    \subfloat[CEW dataset augmented samples after applying FGSM ]{\includegraphics[width=0.7\columnwidth, scale=0.4]{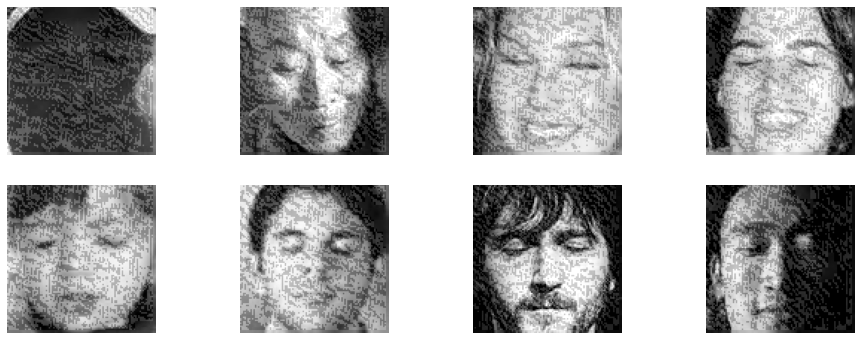}}
    \qquad
    \subfloat[CEW dataset augmented samples after applying DeepFool ]{\includegraphics[width=0.7\columnwidth, scale=0.4]{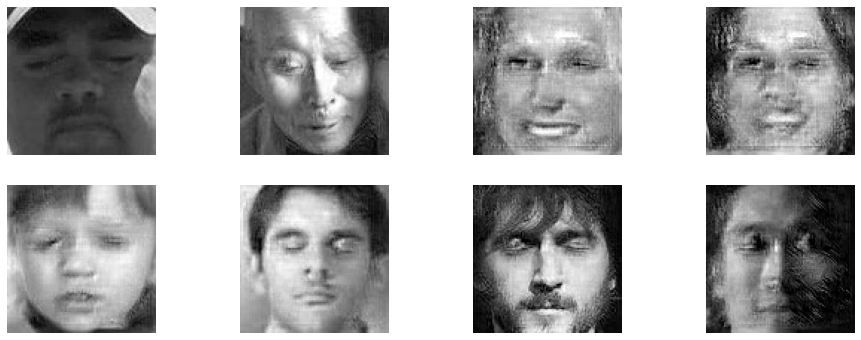}}
    \caption{Baseline images, FGSM, PGD and DeepFool adverserial attacks on augmented Closed Eyes in the Wild dataset.}
    \label{fig:cew_with_augmentation}
\end{figure}

\end{document}